\pdfoutput=1

\documentclass[11pt]{article}

\usepackage[]{acl}

\usepackage{times}
\usepackage{latexsym}
\usepackage{mathtools}
\usepackage[T1]{fontenc}

\usepackage[utf8]{inputenc}
\usepackage{amsmath, amssymb}
\usepackage{microtype}
\usepackage{bbm}
\usepackage{url}
\usepackage{changes}    
\usepackage{booktabs}
\usepackage{physics}
\usepackage{xcolor}

\newcommand{\blockcomment}[1]{}

\usepackage{nicefrac}

\usepackage{mathtools, bm}

\usepackage{xspace}
\newcommand{\ie}{\emph{i.e.}\xspace}
\newcommand{\eg}{\emph{e.g.}\xspace}
\newcommand{\etc}{\emph{etc}\xspace}
\newcommand{\ia}{\emph{inter alia}\xspace}

\usepackage[noabbrev,capitalize,nameinlink]{cleveref}

\usepackage{alphabeta}

\usepackage{twemojis}

\title{Uncertainty in Natural Language Generation: \\From Theory to Applications}

\newcommand{\amsterdam}[0]{$^{1}$}
\newcommand{\darmstadt}[0]{$^{2}$}
\newcommand{\itu}[0]{$^{3}$}
\newcommand{\aipc}[0]{$^4$}
\newcommand{\lisbon}[0]{$^5$}
\newcommand{\lisbonist}[0]{$^6$}
\newcommand{\lmu}[0]{$^{9}$}
\newcommand{\zurich}[0]{$^7$}
\newcommand{\edinburgh}[0]{$^8$}

\author{
Joris Baan\amsterdam \thanks{\;\;Equal contributions. Corresp. to \url{j.s.baan@uva.nl}} \ 
Nico Daheim\darmstadt \footnotemark[1] \  
Evgenia Ilia\amsterdam \footnotemark[1] \ 
Dennis Ulmer\itu$^,$\aipc \footnotemark[1] \  
{\bf Haau-Sing Li}\darmstadt \\
{\bf Raquel Fern\'{a}ndez}\amsterdam 
{\bf Barbara Plank}\lmu$^,$\itu%
{\bf Rico Sennrich}\zurich$^,$\edinburgh 
{\bf Chrysoula Zerva}\lisbon$^,$\lisbonist \ 
{\bf Wilker Aziz}\amsterdam
\\
\amsterdam University of Amsterdam
\darmstadt TU Darmstadt \& hessian.AI
\itu IT University of Copenhagen \\
\aipc Pioneer Centre for Artificial Intelligence
\lisbon Instituto de Telecomunica\c{c}\~{o}es\\ 
\lisbonist Instituto Superior Técnico \& LUMLIS (Lisbon ELLIS Unit)
\zurich University of Zurich\\
\edinburgh University of Edinburgh
\lmu LMU Munich \& Munich Center for Machine Learning
}
\begin{document}

\maketitle

\begin{abstract}

Recent advances of powerful Language Models have allowed Natural Language Generation (NLG) to emerge as an important technology that can not only perform traditional tasks like summarisation or translation, but also serve as a natural language interface to a variety of applications. As such, it is crucial that NLG systems are trustworthy and reliable, for example by indicating when they are likely to be wrong; and supporting multiple views, backgrounds and writing styles---reflecting diverse human sub-populations.
In this paper, we argue that a principled treatment of \textit{uncertainty} can assist in creating systems and evaluation protocols better aligned with these goals.
We first present the fundamental theory, frameworks and vocabulary required to represent uncertainty. We then characterise the main sources of uncertainty in NLG from a linguistic perspective, and propose a two-dimensional taxonomy that is more informative and faithful than the popular aleatoric/epistemic dichotomy. Finally, we move from theory to applications and highlight exciting research directions that exploit uncertainty to power decoding, controllable generation, self-assessment, selective answering, active learning and more.

\end{abstract}

\section{Introduction}

Natural Language Generation (NLG) has long been one of the ultimate goals of artificial intelligence, exemplified by the Turing test \citep{turing1950computing} and systems such as ELIZA \citep{weizenbaum1966eliza} and Watson \citep{high2012era}. It includes a vast number of applications like translation, 
summarisation, question answering, story telling and image captioning. Recently, NLG systems are gaining traction as general purpose interfaces through which users can interact with any application using natural language (\eg, Google's Bard, OpenAI's ChatGPT and Meta's LLaMA; \citealp{google_bard, openai_chatgpt, touvron2023llama, touvron2023llama2}). Their widespread use makes it increasingly important to build NLG systems that are trustworthy and representative of the diversity of its users \citep{bhatt-uncertainty-2021,jacovi2021formalizing, liao2023ai}. 

In this paper, we argue that an explicit and principled treatment of \textit{uncertainty} can assist in creating systems better aligned with these goals. For example by enriching predictions with a degree of confidence that is predictive of error, and ensuring that multiple views, backgrounds and writing styles from diverse populations are well represented. Practical applications of high quality, and, importantly, \textit{disentangled} representations of uncertainty include the ability to abstain from answering, ask clarification questions, or defer a decision; to predict a distribution over a range of human interpretations and perspectives, or to cater to specific users through controllable generation.

One key aspect of human language that makes a principled treatment of uncertainty in NLG models particularly important and challenging is that given a context---for example, a dialogue history---there are many different ways in which someone might respond  (\eg, \citealp{scovel1998psycholinguistics}, p. 37). From the perspective of an NLG system, this causes \textit{uncertainty} about the desired response \citep{ferreira2000effect}. Similar observations have been made in classification, where annotators might pick a different but equally plausible class label given a task and an input text---a line of work recently united under the term \textit{human label variation} \cite{plank-2022-problem}. In NLG, however, the unbounded output space (all possible strings) and enormously open-ended tasks make plausible response variability particularly high \cite{giulianelli2023comes}. On top of uncertainty due to rich variability in language use, other sources of uncertainty stem from lack of knowledge about the plausibility of the NLG system itself, for example, due to neural architecture or parameter setting. Most NLG models, however, have no disentangled representation---or no representation at all---for these sources of uncertainty. 

Despite its importance, (survey) papers on uncertainty in machine learning do not discuss NLG \cite{abdar-review-2020, gawlikowski-survey-2021, hullermeier2021aleatoric, mena-survey-2022, gruber2023sources}. Likewise, those about NLG do not discuss uncertainty \cite{gatt-survey-2018, celikyilmaz-evaluation-2020, hi-survey-2022, erdem-neural-2022, dong-survey-2022}. One exception is concurrent work by \citet{hu2023uncertainty} from which we differ by presenting a comprehensive theoretical background and identifying the main sources of uncertainty in NLG from a linguistic perspective, with complementary applications.

We summarise our contributions as follows. To encourage a principled and shared understanding of uncertainty, we provide a detailed account of uncertainty, and how to represent, learn and reason about it in \Cref{sec:background}. We characterise the main sources of uncertainty in NLG from a linguistic perspective by proposing the \textit{double triangle of language production}, and a fluid, two-dimensional taxonomy that is more informative and faithful than the popular aleatoric/epistemic dichotomy in \Cref{sec:sources}. Finally, we move from theory to exciting applications of (disentangled) representations of uncertainty, such as decoding, controllable generation, self-assessment, selective answering, and active learning in \Cref{sec:applications}.
\section{Uncertainty}
\label{sec:background}
Uncertainty is a rich concept that has received various reasonable treatments before today's understanding of it.\footnote{From uncertainty's connection to (mostly abandoned) views on what is `knowable' \citep{knight1921risk}, to its central role in decision theories  \citep{ramsey1931foundations,von1947theory,Wald1951StatisticalDF,bernardo1994bayesian} and mathematical statistics \citep{savage1972foundations} to its modern understanding in terms of state of knowledge \citep{morgan_henrion_1990,lindley2013understanding}, to its  mathematical representation detached from philosophical interpretation \citep{halpern2017reasoning}.} We begin discussing it through common language. The online edition of the Oxford English dictionary listed five senses of uncertainty (retrieved in May 2023), two of which we partly quote here (those general enough to include the others as special case): \emph{(i) the state of not being definitely known or perfectly clear}; and \emph{(ii) the state or character of being uncertain in mind}. Both definitions regard uncertainty as \emph{a state of affairs}: in \emph{(i)}, the state of the world; in \emph{(ii)}, the state of an agent contemplating the world. They are subtly different: \emph{(i)} encompasses situations of inherent randomness (\eg, the result of a coin flip), \emph{(ii)} concerns one's inability to predict the state of the world regardless of any inherent randomness (\eg, a reader wondering about the content of the next paragraph). As we shall see, this difference leads to rather different interpretations %
of uncertainty as an aspect of reality. Yet, at the level of mathematical treatment, they share the same formal devices. Hence, with no loss of generality, we choose to talk about uncertainty from the point of view of an agent contemplating or interacting with the world, while possessing limited knowledge about it. Our presentation is inspired by various reference texts,  in particular, \citet{dubois2009formal} and \citet{halpern2017reasoning}. 

\paragraph{Agents.} We posit that any one agent shall represent the state of their knowledge in a way sufficient for reasoning about the truth value of claims (or propositions) that they make about aspects of the world. In particular, the agent is able to state their preference for claims they find themselves less uncertain about (\ie, possessing better information about those).\footnote{Agents and worlds are abstractions to be adapted and tailored to each application, commonly in NLG an agent is a model and a world is a response to a given prompt.}  An agent then uses this \emph{uncertainty representation} to interact with the environment (\eg, inform their actions) and, when they acquire new knowledge, they update the representation in a coherent manner. 
To illustrate formal concepts, we use three example agents. \textbf{A1}\hspace{0.5mm}\twemoji{game die} rolls a six-sided die; we seek to represent their state of knowledge about the outcome. \textbf{A2}\hspace{0.5mm}\twemoji{busts in silhouette} resolves mentions of entities to unique names in a knowledge base (KB); we seek to represent their state of knowledge about entity names given any one mention. Last, \textbf{A3}\hspace{0.5mm}\twemoji{speech balloon} provides written answers to questions; we seek to represent their state of knowledge about answers given any one question. 
For simplicity, we assume that our agents already acquired their knowledge, by means which are not relevant for now, and their state of knowledge is frozen.  
We begin by outlining the formal tools common to all frameworks for uncertainty representation we are aware of, we then zoom into the most commonly used framework (probability) and discuss the role of statistics in acquisition and revision of knowledge.  

\paragraph{Possible worlds.} Our agent does not know the state of the actual world, but they assume that it must be one of a collection of possible worlds (the universe). They represent a world as a unique symbol (or string, or collection of attributes; the level of detail being dictated by the agent's needs), and the universe of what is possible as a set $\Omega$ of mutually exclusive worlds.\footnote{This framework, \textit{possible worlds}, is familiar to linguists and philosophers alike \citep{hintikka1957modality,hintikka1961modality,sep-possible-worlds}.
} 
\textbf{A1}\hspace{0.5mm}\twemoji{game die}  might represent a world as a symbol $f_k$, with $k$ denoting the number of pips the die shows as a result of the roll; they might assume the die always lands showing one of six numbered faces and thus take $\{f_1, \ldots, f_6\}$ to represent all possible worlds. 
For \textbf{A2}\hspace{0.5mm}\twemoji{busts in silhouette}, a world is a symbol like $e_i$, with $i$ denoting an entity's identifier (\eg, a standardised unique name), and the universe is the finite set of entities in the English Wikipedia. 
For \textbf{A3}\hspace{0.5mm}\twemoji{speech balloon}, a world is a symbol like $u_s$, with $s$ an English sentence produced in response to a question. This agent happens to be unable to describe the set of all valid English sentences (they cannot enumerate its elements nor state a finite set of properties that all valid sentences must satisfy). Motivated by convenience, \textbf{A3}\hspace{0.5mm}\twemoji{speech balloon} uses a set large enough to encompass most of it while being specifiable in a compact manner: the set of all finite-length strings made by concatenation of known symbols (\eg, words, punctuation, \etc). These examples show that the agent's choice of universe can be a difficult one, often requiring simplifying assumptions: on soft or irregular terrain, a die could land on an edge; a KB may be incomplete (sometimes in known ways, \eg, under-representing the contributions of Black women to science); a regular language is a too loosely constrained representation of the English language (\eg, it includes infinitely many strings that will never correspond to any actual world). 

\paragraph{Propositions.} The possible worlds framework gives agents a mechanism to represent claims about specific aspects of the world. A \emph{proposition}  $E_t$ is the claim that the actual world $\omega$ is one where some property $t$ holds (which we denote $t(\omega)$). A property is something that can be assessed for any one world (\eg, $f_2$ is even and prime, $e_{\operatorname{Katherine\_Johnson}}$ is African-American, female and mathematician, $u_\text{`Biden is the 46th US president'}$ expresses the relation $\operatorname{presidentof}(\operatorname{Joe\_Biden}, \operatorname{USA})$). Not knowing the state of the actual world,  our agent represents $E_t$ by the set  $E_t = \{w \in \Omega: t(w)\} \subseteq \Omega$ of all possible worlds where the property holds. If the agent knew the state of the actual world $\omega$, then the truth value of the proposition would be determined simply by set membership (\ie, $\omega \in E_t$ or $\omega \not\in E_t$). 
For example, \textbf{A1}\hspace{0.5mm}\twemoji{game die}  represents the claim ``the roll is odd'' as $E_{\text{odd}} = \{f_1, f_3, f_5\}$.
\textbf{A2}\hspace{0.5mm}\twemoji{busts in silhouette} represents the claim ``mention to a female mathematician'' by the set $\{e_i \in \Omega: \operatorname{female}(e_i) \wedge \operatorname{mathematician}(e_i)\}$. %
\textbf{A3}\hspace{0.5mm}\twemoji{speech balloon} might use equivalence classes, for example, they use the set $E_a = \{u_s \in \Omega : \operatorname{equivalent}_{a}(u_s)\}$ to claim that the answer is a sentence semantically equivalent to some other sentence $u_a \in \Omega$ (\eg, $u_\text{`The 46th US president is Joe Biden'}$). Because propositions are semantic in nature, they can be difficult to represent explicitly. For example, \textbf{A3}\hspace{0.5mm}\twemoji{speech balloon}'s equivalence classes require sophisticated natural language understanding. 
A representation of all propositions an agent deems possible is a set $\mathcal E$ of subsets of $\Omega$.\footnote{If an agent has no knowledge of the impossibility of any proposition, or does not care to exclude those from the representation, the powerset of (countable) $\Omega$ is a reasonable choice for $\mathcal E$. In NLG, we often implicitly make this choice.}

\paragraph{Preferences.} The agent's imperfect knowledge of the actual world $\omega$ translates to limited knowledge about propositions. However, the agent's ignorance is qualitatively different depending on the claims they make. Intuitively, some claims are compatible with many of the possible worlds, while others hold in but a few (\eg, \textbf{A1}\hspace{0.5mm}\twemoji{game die}  knows that only one prime number is even), and though the various worlds are all possible, they may not be equally plausible (\eg, \textbf{A2}\hspace{0.5mm}\twemoji{busts in silhouette} knows that most mentions resolve to politicians, \textbf{A3}\hspace{0.5mm}\twemoji{speech balloon} knows that most answers are only a few words long), \etc.
Considerations of those kinds motivate an agent to express a \emph{preference} for claims they find themselves less uncertain about (\ie, possessing better information about those). The agent does so by prescribing a \emph{plausibility measure} \citep{friedman96}, a function that attaches a token of uncertainty---a qualifier that the agent knows how to sort---to each proposition in $\mathcal E$. Plausibility measures are very diverse, the most well known instance of it being axiomatic probability \citep{kolmogorov1960foundations}.\footnote{Other plausibility measures include belief functions \citep{shafer76}, possibility measures \citep{duboisprade90}, ordinal ranking functions \citep{GoldszmidtPearl92} and (non-numerical) preference orders \citep{friedman96}. 
Concrete instances of plausibility measures vary in descriptive power. Under certain documented assumptions \citep{friedman96}, they enable something like a `calculus of uncertainty' which formalises the procedures the agent must follow to incorporate additional information about the world and revise their uncertainty representation coherently (in axiomatic probability, this is known as \emph{conditioning}).} 

\paragraph{Probability.} Probability is a numerical qualifier that we can attach to events in $\mathcal E$.\footnote{In probability, propositions are  \emph{events}, worlds are \emph{outcomes} and universes are \emph{sample spaces}.} 
For any event, this qualifier is a positive real number bounded to be at most $1$. Probability values inherit various properties of real numbers: we can add, multiply and sort them. A function $\Pr$ over $\mathcal E$ is a \emph{probability measure} if a) it maps $\Omega$ to $1$, and b) it maps any two disjoint sets $U$ and $V$ in $\mathcal E$ to $\Pr(U)+\Pr(V)$, which is known as additivity. 
 Additivity, in particular, implies that we can identify a probability measure over all of $\mathcal E$ by assigning probability to elementary outcomes in $\Omega$, for example, using a probability mass function (pmf) or probability density function (pdf; for uncountable $\Omega$). This has massive consequences for uncertainty representation, since working with elementary outcomes is much simpler than working with sets of outcomes (for example, difficulty in prescribing equivalence classes such as `all sentences that talk about Joe Biden' need not stop \textbf{A3}\hspace{0.5mm}\twemoji{speech balloon} from identifying a probability measure for their reasoning needs).
 
\paragraph{Interpretations.} 
Probability has been motivated and justified from different angles, each building on a specific interpretation of what probability as a number must signify \citep{hacking1975emergence}. However different they are, they all lead to the same formal device. Under certain idealisations, \textit{objectivists} regard events as \emph{repeatable} (\eg, we may prompt a human speaker multiple times). Repetitions allow an agent to perceive what may be thought of as an inherent \emph{property} of an event: its \emph{frequency} in a large enough number of repetitions. The \emph{subjectivist} interpretation \citep{ramsey1931foundations,definetti2017theory} views probability as a degree of belief, personal to an agent, and deprived of any interpretation beyond its formal role as an expression of the agent's preferences.\footnote{Dictionary definition \textit{(i)} is objectivist; \textit{(ii)} subjectivist.} 
Different interpretations have an impact on the procedures that an agent acknowledges as logical or rational for knowledge acquisition and revision, as we discuss next.

\paragraph{Statistics.} We have described the general tools that agents can use to represent and convey their uncertainty. But where do their preferences (probabilities, in particular) come from? The \emph{Frequentist} agent is essentially an objectivist who assumes the existence of a precise statistical law that describes the phenomena in consideration. They assume to have access to this law up to an unknown parameter $\theta^\star \in  \mathbb R^D$. %
Given a parameter $\theta$, their preferences are specified via a pmf (or pdf) $p(x|\theta)$. Given data $\mathbf x = \langle x_1, \ldots, x_N \rangle$, this law identifies the so called likelihood function $\ell_{\mathbf x}(\theta) = \prod_n p(x_n|\theta)$, a measure of the compatibility between observed data $\mathbf x$ and the statistical model identified by $\theta$. The agent uses $\mathbf x$ to estimate the parameter $\theta^\star$: they pick the parameter $\hat\theta$ that maximises the likelihood function, %
this is known as maximum likelihood estimation (MLE). They do not entertain parameters as part of the possible worlds, hence have no uncertainty representation about them. Given the  parameter  estimate $\hat\theta$, the agent uses $p(x_{n+1}|\hat\theta)$ to make predictive inferences about future data $x_{n+1}$. When necessary (\eg, the agent suspects to have found a better statistical law), the agent studies properties of their parameter estimator(s) by repeated experimentation, for example to establish confidence intervals %
 and other tools for model selection (see for example \citealp{lehmann1993fisher}).
The \emph{Bayesian} agent, a subjectivist, %
also picks a statistical law, but makes no assumption about its correctness. Given some data $\mathbf x$, they too construct a likelihood function $\ell_{\mathbf x}(\theta)$, but use it differently. As a formal tool, probability comes with a mechanism for belief revision: conditioning. %
To make use of it, the agent augments their possible worlds to include possible values of $\theta$ and its interaction with possible values of the observable variable, they then state their preferences over parameters in the form of a pdf $p(\theta)$. %
This is called a \emph{prior} (conveys one's knowledge and experience before observing $\mathbf x$).
The agent then revises their preferences using Bayes rule %
to obtain a posterior pdf $p(\theta|\mathbf x) \propto p(\theta)\ell_{\mathbf x}(\theta)$. This object supports all inferences the agent will ever make (\eg, 
about parameters, or about future data $x_{n+1}$---for which the agent builds a posterior predictive function  $p(x_{n+1}|\mathbf x) = \int p(x_{n+1}|\theta)p(\theta|\mathbf x) \dd{\theta}$).
In essence, Frequentist procedures revolve around point estimation (\eg, MLE) and null hypothesis significance testing \citep{LehmCase98,lehmann2005testing}, %
Bayesian theory \citep{bernardo1994bayesian} and practice \citep{gelmanbda04}, instead, frame statistical inference as an iterative process of belief revision  (\eg, conditioning, marginalisation, expectation).

\paragraph{Natural Language Generation.} 
Most NLG models (like \textbf{A3}\hspace{0.5mm}\twemoji{speech balloon}) acquire knowledge through MLE. Alternatives include Bayesian inference \cite[\eg,][]{malinin2020uncertainty,sankararaman2022bayesformer} and utility- and reward-based training (\eg, minimum risk \citep{shen-etal-2016-minimum},  reinforcement learning \citep{paulus2018a}). Recently, pre-training on enormous unlabelled corpora, and reinforcement learning from human feedback \cite[RLHF, \eg,][]{christiano2017deep,stiennon2020learning,ouyang2022training} or \textit{instruction tuning} \cite[\eg,][]{mishra-etal-2022-cross,wei2022finetuned} have become popular to refine the representation of uncertainty towards something that decodes more easily into strings preferred by human users. %

Generating a response is simulating an outcome. %
The event space is the powerset of all token sequences from a fixed vocabulary \cite[BPE tokens, \eg,][]{sennrich-etal-2016-neural}. Rather than prescribing a probability measure (mapping each event to a probability value) directly, we parameterise a pmf (typically parameterised via an autoregressive factorisation of the probability of any one sequence) with a neural network and exploit countable additivity to assign probability to any event (\eg, all token sequences that map to the same sentence \citep{cao-rimell-2021-evaluate,chirkova-etal-2023-marginalize} or all sentences that map to the same meaning representation \cite{kuhn2023semantic}). %

\paragraph{Key Takeaways.}
(1) Uncertainty is a state to be represented. %
(2) To represent uncertainty about something observable or not (\eg, responses, parameters, modelling assumptions) we need to acknowledge and order a whole space of alternatives: our choice of possible worlds must capture interaction amongst possible values of the variables we aim to express our uncertainty about. 
(3) Probability is not constrained to abide by any one interpretation. To regard probabilities in a specific human-interpretable way (\eg, relative frequencies), we need learning techniques yielding that result, and we need to verify that our setting actually meets all necessary conditions (\eg, the Frequentist interpretation of probability is sensitive to modelling choices, local optimality, and  data sparsity: most practical NLG agents are unable to meet the necessary formal requirements).
\section{Sources of Uncertainty in NLG}
\label{sec:sources}
From the perspective of an NLG agent, there are many phenomena or \textit{sources} that lead to uncertainty about a response given a context. A \textit{disentangled} representation of these sources is crucial. For example, to ask a clarification question when the context is under-specified or ambiguous or to abstain from answering when the context is not familiar to the agent (more applications in \Cref{sec:informed_decisions}). In this section, we identify and characterise the main sources of uncertainty when generating natural language. 
We start by characterising sources of uncertainty in human language production; the data.\footnote{Data might nowadays include generated text, see \citet{taori2022data, veselovsky2023artificial}.} 
We then move to sources of uncertainty in modelling decisions, and finally propose a fluid, two-dimensional taxonomy that we argue to be more informative and faithful than the popular aleatoric/epistemic dichotomy.

\subsection{Sources of Uncertainty in Data}
\label{sec:uncertainty-data}
As discussed in \Cref{sec:background}, multiple possible worlds give rise to uncertainty. Variation in human language production, therefore, gives rise to uncertainty about a response given a context. While sources of disagreement in classification tasks have been studied extensively \cite{poesio-etal-2019-crowdsourced,basile-etal-2021-need,plank-2022-problem,jiang-tacl_a_00523,sandri-etal-2023-dont}, there is less understanding of where variation in NLG data stems from. 

\begin{figure}
    \begin{center}
        \includegraphics[width=.95\columnwidth]{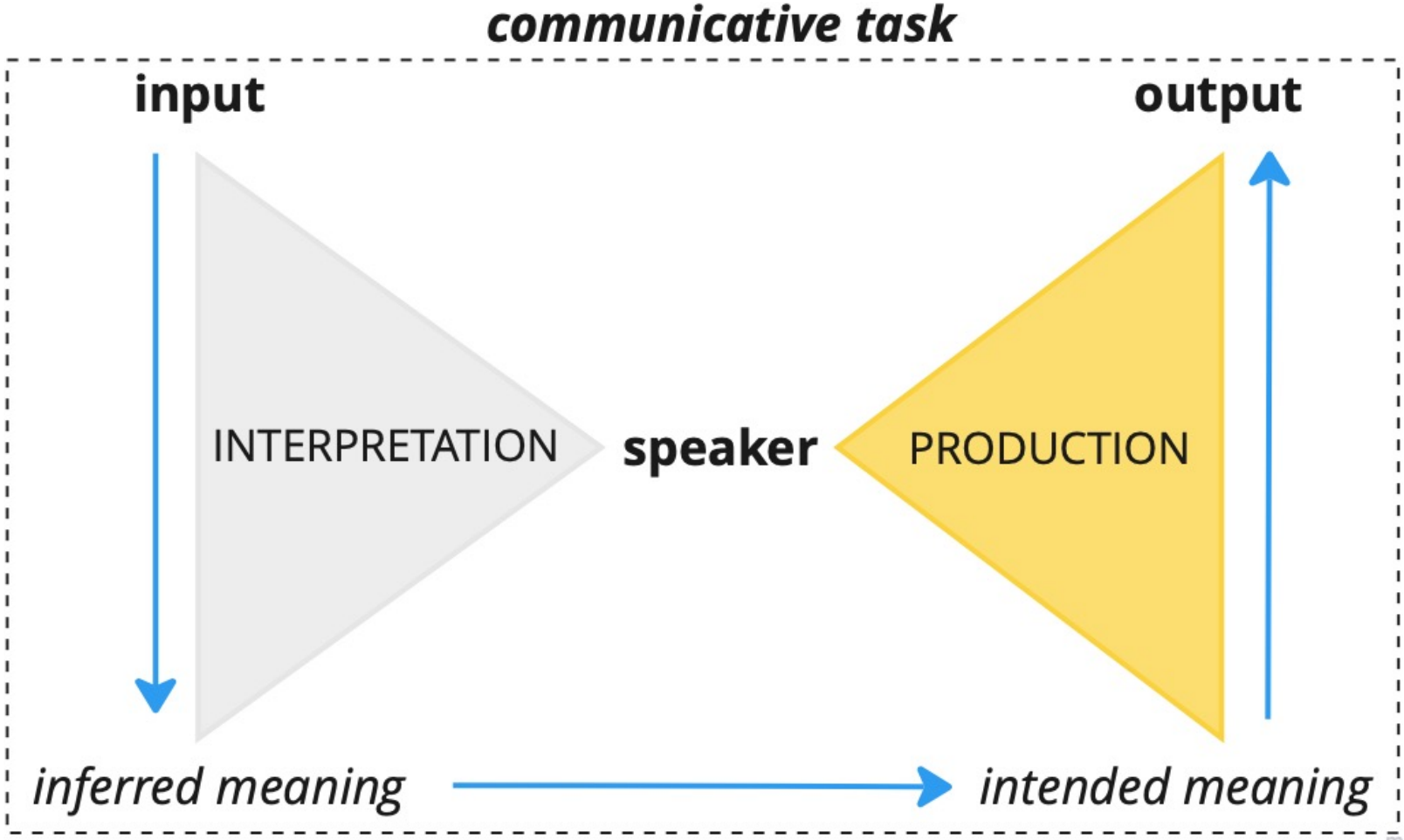}
        \caption{We propose the \textit{double triangle of language production}, extending the traditional ``Triangle of Reference'' \cite{ogden-richards-1923meaning} to NLG: a schematic diagram of the key aspects involved when a speaker produces a linguistic signal (output) given a prompt (input) in the context of a communicative task. The blue arrows correspond to sub-processes where one-to-many mappings may arise, thus leading to possible variation, which gives rise to uncertainty.}
        \label{fig:production-diagram}
    \end{center}

\end{figure}

To address this issue, we build on the Triangle of Reference \cite{ogden-richards-1923meaning}, used to identify types of disagreement in classification \cite{Aroyo_Welty_2015,jiang-tacl_a_00523}, and extend it to the more complex case of language production. 
\cref{fig:production-diagram} shows a schematic diagram of the key aspects involved in the production of a linguistic signal (output) by a speaker given a prompt (input; be it a sentence in a source language, an image, a text, or a sequence of dialogue turns, \ia) in the context of a communicative task, such as translation, image description, summarisation, or dialogue response generation. 

The left-hand-side triangle in this diagram is akin to the ``Triangle of Reference'', which here corresponds to the \textbf{input interpretation process}: the speaker infers a meaning for the input. Such meaning is not explicitly constrained and remains unobserved (unlike in classification tasks, where the space of possible meanings is restricted by the possible labels an annotator can assign). The right-hand-side triangle depicts the \textbf{output production process}: the interpretation of the input in the context of a given task evokes a communicative goal or intended meaning (\textit{what} the speaker wishes to convey), which is then overtly expressed by the output (\textit{how} the intent is conveyed via language).\footnote{These two latter steps loosely correspond to content determination and surface realisation in traditional data-to-text NLG \citep{reiter_dale_2000,gatt-survey-2018}.}

We are interested in characterising the sources of variation in the output for a fixed input.\footnote{Typically this will correspond to variability across speakers (\eg, different summaries of a given text produced by different crowdworkers), but nothing hinges on this: different outputs may be produced by the same individual.}  
We introduce these sources in the context of different production sub-processes where a one-to-many mapping can occur. The blue arrows in \cref{fig:production-diagram} refer to such sub-processes.\footnote{While in this section we focus on sources inherent to the language production process, data collection design can also have an impact on output variation \cite{geva-etal-2019-modeling, parmar-etal-2023-dont} (\eg, 
insufficient guidelines regarding the target length of a summary, the use of slang, or the level of detail with which an image should be described). Yet, many current NLG systems are not trained on curated datasets with annotator guidelines but on fortuitous data such as movie subtitles or online user-generated content.}

\begin{figure*}
    \centering
    \includegraphics[width=\textwidth]{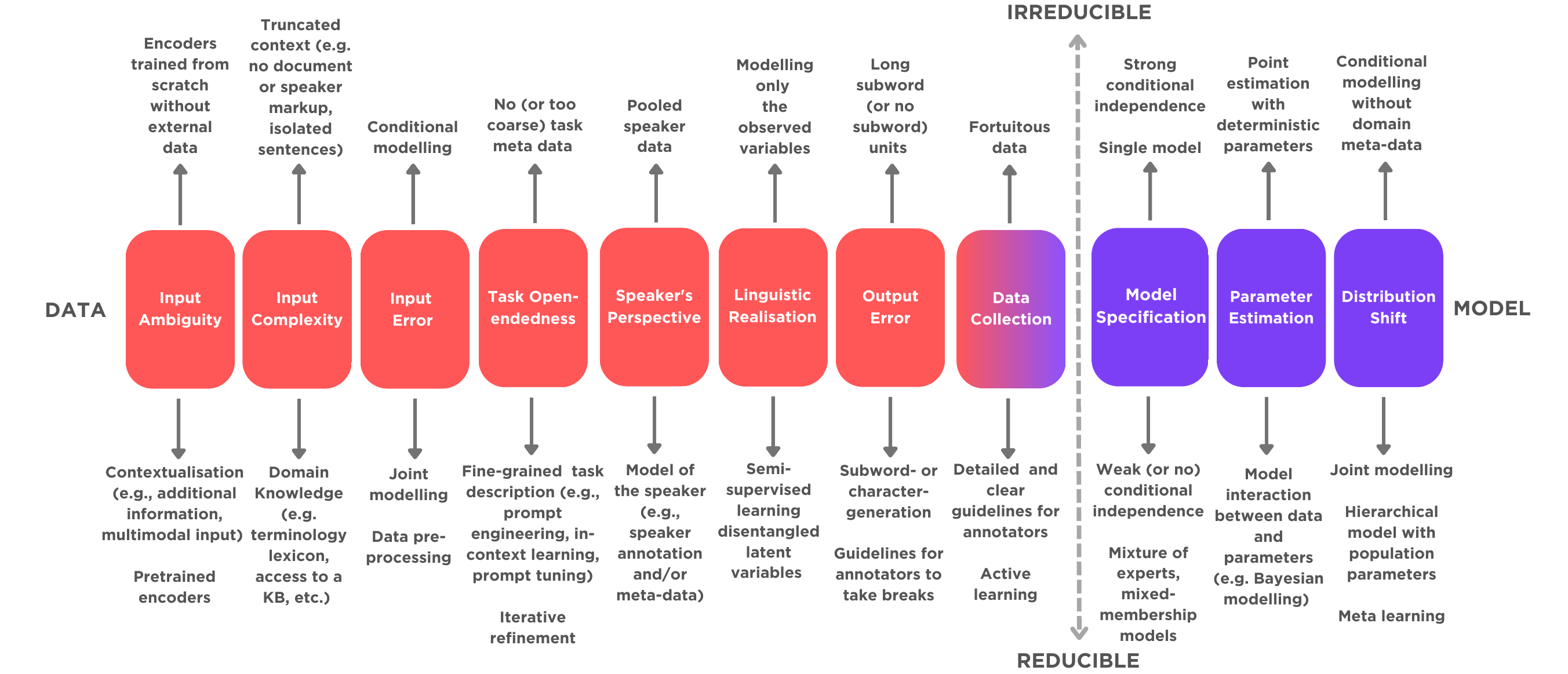}
    \caption{The main \emph{sources of uncertainty} in NLG relate to the nature of the data and the tools available to the modeller. We depict them in red and blue (data and model respectively). Being able to reduce uncertainty is hardly determined by the sources themselves, rather, reducibility depends on decisions (from data collection to model design and training) by the NLG practitioner. For each source, we list along the vertical axis some considerations that affect reducibility. This view generalises the traditional aleatoric vs. epsitemic categorisation of uncertainty. 
    }
    \label{fig:spectrum}
\end{figure*}

\paragraph{Input {\color[HTML]{4d99df}$\rightarrow$} inferred meaning.} 
Variation in the output can stem from multiple ways to infer a meaning for the input. %
Such variation has three main sources: 
\textit{input ambiguity}, 
\textit{input complexity}, and 
\textit{input error}. 
Input ambiguity can be due to polysemous words (\eg,``I have a date'', where ``date'' may be a fruit, an appointment or a certain day), syntactic ambiguity (\eg, ``the fish is ready to eat'', where the fish may be ready to be fed or to be eaten) or underspecification (\eg, in pro-drop languages where the subject, and hence its grammatical gender, is not explicit:  in Greek, `Tρώει.' may be translated as ``she/he/it eats.'').   
The degree of complexity in the input---such as the   presence of many entities in an image or of infrequent words in textual prompts---may result in different degrees of processing difficulty and thus also lead to variation in the interpretation process. 
Finally, input error (\eg, spelling errors) may also lead to noise in how speakers interpret the input and consequently to output variation. 

\paragraph{Inferred meaning {\color[HTML]{4d99df}$\rightarrow$} intended meaning.} 
Given a fixed input interpretation, outputs with different intended meanings may be plausible, thus leading to semantic variation in the output. 
This is modulated by two main factors: the \textit{open-endedness} of the task and the \textit{personal perspective} of the speakers. 
For example, in more open-ended tasks, such as dialogue response generation or storytelling, a given prompt may be followed up by semantically diverse dialogue acts or story continuations. This contrasts with tasks such as machine translation, where the intended meaning behind the output is determined by the input's meaning. In addition, speakers may have different perspectives due to different social backgrounds or points of view. Thus, if a task allows for semantic variation, such diversity of perspectives is likely to result in diverse outputs. For example, asking different individuals to answer the question `Should we have guns at home?' might trigger a range of different intents based on geographical and societal backgrounds.

\paragraph{Intended meaning {\color[HTML]{4d99df}$\rightarrow$} output.}
The sources of variation discussed above lead to semantic variation (\ie, different surface-form outputs with different meanings). 
However, given a fixed inferred meaning for the input and a fixed intended meaning, we may still observe variation in the output (\ie, different surface-form outputs that are semantic paraphrases). There are indeed many ways to express the same idea. Such meaning-preserving  variation stems from the \textit{non-determinism} of the cognitive processes involved in linguistic realisation \cite{levelt1993speaking}. For example, speakers' cognitive and social differences may lead to variability related to lexical and construction accessibility: \eg, choosing among different synonyms or between an active vs.\ a passive syntactic structure. Similarly, the cognitive effort of linguistic production itself may result in less than optimal outcomes including \textit{output errors} \cite{goldberg2022good}, which again are likely to lead to variation.

\subsection{Sources of Uncertainty in Modelling Decisions}
\label{sec:uncertainty-modelling-decision}
There are many ways to model language production. These decisions lead to additional uncertainty about the plausibility of a generation. As discussed in \Cref{sec:background}, to represent, learn and reason about uncertainty, these modelling decisions must be included in the possible worlds. 
We identify three main sources of model uncertainty.

\paragraph{Model Specification.}
Uncertainty about decisions that influence the hypothesis space, including neural architectures such as attention, enforced sparsity, parameter tying, number of parameters, decoding algorithm, width, depth, factorising sequence-level probability, etc.

\paragraph{Parameter Estimation.}
Uncertainty about parameters given a specified model. This includes decisions about optimisation objective, learning algorithm, training time (epochs), batch size, initialisation, etc.

\paragraph{Distribution Shift.}
Uncertainty due to a shift in distribution from observed data points to unseen data points. Shifts can be co-variate, where the input-output mapping remains constant but the input distribution shifts, or conditional, where the input distribution shifts. More details in \citep[\eg,][]{moreno2012unifying, hupkes2022state}.

\subsection{Aleatoric vs.\ Epistemic Uncertainty}
\label{sec:aleatoric-epistemic}
\textit{Aleatoric} and \textit{epistemic} uncertainty are popular terms in the deep learning community, and often used to categorise sources of uncertainty. Their precise definition, however, is not obvious \cite{der2009aleatory, hullermeier2021aleatoric}. Aleatoric uncertainty is often described as irreducible and related to data; epistemic as reducible and related to a model. We take the stance that this dichotomy \textbf{conflates two aspects} that are arguably orthogonal for most (if not all) sources of uncertainty. %
Namely, the agent's uncertainty being qualitatively different as a function of the \textbf{source} (\ie, the data being modelled or the model being used) and its \textbf{reducibility}. %
Reducibility depends on the ability and willingness of the practitioner to represent, learn and reduce a source of uncertainty, and, while the pathways to do so depend on the precise source, whether the source is related to data or model seems of little to no importance. 
For example, if uncertainty about parameters---clearly related to model  and often considered reducible---is not represented (\ie, not part of possible worlds), it cannot be reduced (unlike the common notion of epistemic uncertainty suggests); if annotator guidelines influence human response variability---clearly related to data and often considered irreducible---it \textit{can} be reduced (unlike the common notion of aleatoric uncertainty suggests).\footnote{Our view resonates with recent views in statistics: in concurrent work, \citet{gruber2023sources} echo that reducibility can only be discussed in the context of a choice of data and model, advocating examining different \emph{sources} of uncertainty.}

We propose a taxonomy beyond aleatoric/epistemic in \Cref{fig:spectrum} that depicts two dimensions. To the left we have sources that are linked to what the agent contemplates or interacts with (under the label `data'). To the right we have sources that are linked to the tools and assumptions the agent makes (under the label `model'). The boundary between the two is not always clear cut, as some sources can be regarded one way or the other (\eg, input error relates to data, but a model that encodes words at character level will be affected very differently than one that uses word2vec). The vertical axis captures reducibility and we illustrate with labelled arrows  examples of changes the agent can make in order to move a source along the reducibility dimension.
\section{Applications in NLG}
\label{sec:applications}
We have laid down the theory and tools required to \textit{represent}, \textit{learn}, and \textit{reason about} uncertainty in \Cref{sec:background}, and characterised important \textit{sources} of uncertainty in NLG in \Cref{sec:sources}. 
We now highlight how to leverage this to create decision-making systems and evaluation frameworks that are flexible and representative of the diversity and complexity of language and its speakers, and can self-assess their own plausibility. We discuss applications exploiting disentangled representations of uncertainty related to \emph{data} in \Cref{sec:informed_decisions} and 
related to related to a \emph{model} in \Cref{sec:self_assessment}.

\subsection{Exploiting Data Uncertainty}
\label{sec:informed_decisions}
At its core, a good representation of uncertainty allows for informed decision making, weighing potential outcomes (\ie, possible worlds) against the risks or rewards associated with them. In NLG, the predictive distribution over tokens reflects a model's `beliefs' given modelling assumptions, data and learning algorithm. A decision is generating a string given a prompt by composing a sequence of tokens (\ie, \textit{decoding} token-level distributions). %

\subsubsection{Decoding}
A common decoding strategy is to search for the mode; the most-likely response under the statistical model (\eg, using beam search). However, this largely neglects the fact that models define a complex and rich representation of uncertainty. Consider two very different distributions with the same mode; one is mostly concentrated, the other is not. This information is ignored, which has been shown sub-optimal \cite[\eg][]{stahlberg-byrne-2019-nmt,eikema-aziz-2020-map,meister-etal-2020-beam}.
Decoding algorithms can increase diversity, quality, and similarity to the data by better utilising the full representation of uncertainty \cite[\eg][]{fan-etal-2018-hierarchical,Holtzman2020The,hewitt-etal-2022-truncation,li-etal-2016-diversity, zhang2018generating}; power utility-aware objectives \cite[\eg][]{eikema-aziz-2020-map, fernandes-etal-2022-quality, freitag2023epsilon,johnson2023ru};
or account for surface-form variation---for example in generative QA \cite{holtzman-etal-2021-surface}.

\subsubsection{Controllable Generation}
Disentangled representations of uncertainty provide valuable information about and control over types of variability in the data as captured by the model. This can further drive decision making. For example, a user might 1) not care about uncertainty about the syntactic structure of a response, but rather about uncertainty about how to interpret the input, or 2) adapt generations to adhere to a specific style.

\paragraph{Exploiting Latent Context.}
One approach for increased control and interpretability is to introduce a \textit{latent} variable in the possible worlds that represents a specific source of uncertainty. For example, a syntactically ambiguous input causes uncertainty about the response, but this could be decomposed into the uncertainty over structures and the uncertainty of the generation given a specific structure. Past work has attempted to model latent grammatical structures of sequences to guide generations \citep{kim-etal-2019-compound,kim-etal-2019-unsupervised}, for example by matching attention patterns to tree constituents \citep{bradbury2017towards, wang2019tree, ahmed2019improving, nguyen2020tree}. Others model input ambiguity, for example in MT \cite{stahlberg2022scones} or semantic parsing \citep{stengel2023zero}, or employ variational inference \citep{bowman-etal-2016-generating, zhang-etal-2016-variational-neural, calixto-etal-2019-latent, eikema2019auto, brazinskas-etal-2020-unsupervised} for attention mechanisms \citep{DengEtAl2018} and the morphology of the input \citep{Ataman2020A}.

\paragraph{Exploiting Observable Context.}
Sometimes we can exploit an \textit{observable} auxiliary ``context'' $C=c$ that helps reduce uncertainty about the response $Y$ (\eg, a speaker's age, the newspaper an article is published in, \etc). Unlike before we \emph{know} (or predict) $C=c$ and can therefore exploit it.\footnote{This is subtly different from the previous paragraph, where the total uncertainty (marginally across assignments of the auxiliary variable) remains the same---we only \textit{decompose} it and lack the knowledge to actually \textit{reduce} it.}
The realisation $c$ can take many forms, including but not limited to information about the speaker or social context \citep{zhang-etal-2018-personalizing, sennrich-etal-2016-controlling}, knowledge graphs \citep{moon-etal-2019-opendialkg} or unstructured documents (\eg Wikipedia in the case of \citealp{dinan2018wizard}) in dialogue; document-level context in translation \cite{wang-etal-2017-exploiting-cross,zhang-etal-2018-improving,sun-etal-2022-rethinking}. If it is not possible to provide $c$ explicitly, it can also be predicted or retrieved dynamically. 
For example, \citet{lewis2020rag} propose retrieval-augmented generation (RAG) models that are trained to retrieve relevant documents; \citet{yao2022react} propose to train models that execute Wikipedia search calls and condition on them. Later, \citet{schick2023toolformer} train models to execute API calls on other models, calculators, and Wikipedia search, and \citet{gao2022pal} train language models to translate mathematical reasoning tasks into executable programs that generate expected answers. 

Recently, with the rise of LLMs, $c$ often takes the form of \textit{instructions} expressed through natural language \citep{raffel2020exploring}. This powerful mechanism crucially allows us to condition on additional variables (thus revising uncertainty) simply by reformulating the input, without any changes to model design, and has been interpreted as models reasoning over the latent concepts in the context $c$ \citep{xie2022explanation} or being biased by its structure \citep{hahn2023theory}. For instance, steering generations  towards a specific style, length, or content by prompting with initial paragraphs \citep{radford2019language} or providing a detailed description \citep{bubeck2023sparks, zhou2023controlled}.

\subsubsection{Evaluating Decisions under Uncertainty}
\label{subsec:eval-variability}
Plausible response variability is not only challenging for generating high quality text, it also complicates automatic evaluation---especially for metrics that compare individual  generations to individual human responses.\footnote{A single outcome probable under the model is compared to a single outcome recorded in a dataset. Ideally, one compares events: sets of semantically equivalent outcomes, but defining these is non-trivial.} 

BLEU \cite{papineni-etal-2002-bleu}, ROUGE \cite{lin-2004-rouge}, METEOR \cite{banerjee-lavie-2005-meteor} and other word-overlap-based metrics remain de-facto standard for evaluating language generation but fail at capturing the diverse output space, especially for open-ended tasks \cite{gehrmann2022repairing}. Most datasets do not contain multiple references, and generations are rewarded for surface-level similarities to just one reference. Moreover, even datasets that do contain multiple references (observed outcomes) often show poor diversity, even though using multiple high-quality references can greatly improve the reliability of automatic metrics \cite{freitag-etal-2020-bleu}. %
We emphasise their importance when designing evaluation protocols and advocate for collecting multiple references per linguistic context to obtain a sense of plausible variability, as well as establishing datasets that represent different sources of uncertainty. Not observing variability in a dataset does not mean that it does not exist: it is simply a form of data sparsity.

\paragraph{Learned Metrics.}
Learned metrics such as BERTScore \cite{Zhang2020BERTScore}, BLEURT \cite{sellam-etal-2020-bleurt}, or COMET \cite{rei-etal-2020-comet}, have been proposed to overcome the shortcomings of overlap-based metrics but are themselves trained on data that might not reflect sufficient diversity \cite{gehrmann2022repairing}. Therefore, their reliability is heavily dependent upon their ability to generalise, which has been questioned in recent works \citep[inter alia]{amrhein-sennrich-2022-identifying, he2023blind}. Therefore \citet{glushkova-etal-2021-uncertainty-aware} advocate for modelling uncertainty in such metrics too, with \citet{zerva-etal-2022-disentangling} extending it to disentangling different types of uncertainty. %

\paragraph{Statistical Evaluation.}
An interesting evaluation framework complementary to standard NLG evaluation, especially for open-ended tasks, compares \textit{distributions of statistics} in machine generated text to those in human produced text. Usually, this comparison is done globally across an entire test set. For example, by comparing surface-form text statistics like type-token and rank-frequency ratio \cite{meister-cotterell-2021-language}, or by mapping generations and references to an embedding space and comparing the distribution over embedding clusters rather than over strings \cite{pillutla2021mauve,xiang-etal-2021-assessing, pimentel2023on}. 
Some even evaluate generations against human responses \textit{for individual inputs}. For example, by quantifying lexical, syntactic and semantic variability between multiple samples of generation and reference \cite{giulianelli2023comes}; comparing section structure, topics and coreference chains in long-form generation \cite{deng-etal-2022-model}; or comparing sequence length and topic structure \cite{barkhof2022statistical}.

\subsection{Exploiting Model Uncertainty}
\label{sec:self_assessment}
The previous section highlighted applications of disentangled uncertainty about \textit{data} in NLG. Now, we turn towards applications of uncertainty related to \textit{models}. These are often described as mechanisms to ``know what is known'', and revolve around \textit{trustworthiness} of model predictions. Applications of self-assessments include communicating model uncertainty to users to increase trust and transparency \cite{bhatt-uncertainty-2021}, allowing models to selectively answer to reduce errors \cite{el2010foundations}, or improve learning in low-resource settings by selecting data points with high model uncertainty \cite{lewis1995sequential}. Self-assessment in NLG is particularly hard due to the unbounded sample space (all strings); token-level factorised probability; huge models; and high data uncertainty. We discuss several approaches to representing and learning these sources of uncertainty.

\subsubsection{Calibration}
\label{sec:predictive-distribution-and-calibration}
The predictive distribution itself is often taken as a measure of uncertainty indicative of model error, although it is not obvious why this should be the case, as it accounts neither for model specification, nor for parameter estimation as part of the possible worlds. \textit{Calibration} is a frequentist framework that attempts to push probabilities towards the relative frequencies with which predictions are judged to be correct \cite[\textit{inter alia};][]{eisape2020cloze, dhuliawala-etal-2022-calibration, lee-etal-2022-adaptive, zhao2022calibrating}, with various proposed methods such as temperature scaling, label smoothing and knowledge distillation. Recently, \citet{kuhn2023semantic, lin2023generating} show that entropy over clusters of semantically equivalent generations---such a cluster is a good example of an event in NLG---rather than their surface forms is more predictive of model error for QA. These approaches do not disentangle data from model uncertainty, and potentially interfere with good representations of data uncertainty (\Cref{sec:informed_decisions}).

\subsubsection{Conformal Prediction}\label{sec:conformal-prediction}
Another frequentist framework that exploits the predictive distribution 
sorts outcomes by their predicted probability and adds them to a set until some threshold of mass---\eg, 90 \%---is reached.
Though attractive due to its simplicity \citep{kompa2021empirical, ulmer2022exploring}, 
the sets or intervals might be miscalibrated or not contain plausible outcomes. \emph{Conformal prediction} \citep{vovk2005algorithmic, papadopoulos2002inductive, angelopoulos2021gentle} 
uses a held-out calibration set on which the ideal threshold or intervals are computed and guarantees that a plausible prediction will be included with a predefined probability, in expectation. Recently, some works have applied conformal prediction to quality estimation for MT  \citep{giovannotti2023evaluating, zerva2023conformalizing} or even language generation directly \citep{ravfogel2023conformal, quach2023conformal, ren2023robots}.

\subsubsection{Bayesian Inference}

Bayesian methods (\Cref{sec:background} \textbf{Statistics})
offer principled and disentangled representations of uncertainty about the model parameters and the data by entertaining the possibility of multiple parameter settings %
and specifying a probability distribution over them.
Roughly speaking, the spread of the distribution over parameters (or the variance of the generations given a different \textit{sample} of parameters) provides information about parameter uncertainty and can be used to improve active learning \cite{ambati2012active,lyu2020you,gidiotis-tsoumakas-2022-trust}, alleviate hallucinations \cite{xiao2021hallucination}, or provide uncertainty about evaluation metrics \cite{fomicheva2020unsupervised,glushkova-etal-2021-uncertainty-aware, zerva-etal-2022-disentangling}.

In practice, however, the posterior is intractable, motivating approximate methods. For example, based on variational inference \cite{jordan1999introduction}:
\citet{gal2016theoretically} use Monte-Carlo Dropout, \citet{fortunato2017bayesian} extend Bayes-by-Backprop \cite{blundell2015weight} to recurrent models, and \citet{gan2017scalable} use MCMC \cite{robert1999monte}.
\citet{sankararaman2022bayesformer, xiao2020wat} apply MC Dropout to Transformers. 
\citet{zablotskaia2023uncertainty} benchmark methods like Gaussian Process \cite{rasmussen2006gaussian} output layers \cite{liu2020sngp} and how they improve the quality of model uncertainty in summarisation models.
\citet{malinin2020uncertainty} formalise different information-theoretic measures of parameter uncertainty for structured prediction, and in particular their estimators.%

\subsubsection{Verbalised Uncertainty}
Because obtaining probabilities that comply with specific interpretations is difficult (\Cref{sec:background}), an exciting research direction is to re-express the representation with non numerical qualifiers, for example, linguistic expressions such as ``I think'', ``undoubtedly'', or ``high confidence''. \citet{mielke2022reducing} collect human judgements to quantify verbalised uncertainty and quality; measure their correlation; and employ controlled generation to re-generate responses with more aligned verbal expressions of uncertainty. \citet{zhou2023navigating} investigate the effect of adding linguistic (un)certainty markers to prompts, and \citet{lin2022teaching} prompt NLG models to provide a verbal expression of confidence in their response. \citet{kadavath2022language} sample multiple generations and design a prompt asking if one of the proposed answers is true. Then, they assess whether that probability correlates with quality. Although potentially particularly friendly to non-expert users, such re-expressions are not yet well understood, \eg, may not be faithful to the underlying representation of uncertainty they are sampled from.

\subsubsection{Evaluating Self-assessment}
\label{sec:eval_self_unc}
Evaluating the quality of model uncertainty summaries is difficult, because ground truth usually does not exist. Therefore, proxy tasks like error detection \cite{malinin2020uncertainty}, out of distribution detection \citep{malinin2020uncertainty, lahlou2021deup, ulmer2022exploring, van2022benchmarking}, and active learning \cite{osband2023finetuning} are often used.

\paragraph{Error Detection.}
Various NLG studies evaluate how indicative predictive probabilities are of errors or generation quality \cite{jiang2021can, si2023prompting, chen2022close, kumar2019calibration}. This is often operationalised with the Expected Calibration Error \citep[ECE;][]{naeini2015obtaining,guo2017calibration}, which measures whether average predicted probability aligns with model accuracy for groups of predictions. However, ECE's desideratum was recently shown to be unreliable in settings with annotator disagreement as it disregards data uncertainty \citep{baan2022stop}, which might make it ill-suited for NLG. Furthermore %
since evaluating the quality of generations is difficult (\Cref{subsec:eval-variability}), %
notions of correlation between probability and quality suffer from the same issue \cite{jiang2021can, bulian2022tomayto}. Others investigate if predictive probabilities indicate critical errors (such as hallucinations or significant deviations in meaning) in MT \citep{guerreiro2022looking}, abstractive summarisation \citep{van2022mutual}, image captioning \citep{xiao2021hallucination}, and common-sense knowledge \cite{yoshikawa2023selective}. An evaluation framework that is gaining traction is selective answering, \ie, detecting when a model should avoid responding \cite{kamath2020selective, cole2023selectively, kuhn2023semantic, ren2023outofdistribution}, often operationalised with area under the curve of the accuracy/rejection trade-off. 

\paragraph{Out of Distribution Detection.}
The second proxy task is assessing how model uncertainty differs for samples in in-distribution and out-of-distribution settings; essentially performing anomaly detection. The expectation is that models should be uncertain on OOD instances \citep{malinin2020uncertainty, lahlou2021deup, ulmer2022exploring, van2022benchmarking, ren2023outofdistribution}. This task evaluates mostly distribution shift (\Cref{sec:uncertainty-modelling-decision}).

\paragraph{Active Learning.}
Active learning attempts to select the most ``informative'' data instances from which a model will learn the most. The goal is to achieve comparable (or better) performance with less training instances \cite{lewis1995sequential, houlsby2011bayesian, siddhant-lipton-2018-deep,osband2023finetuning}. One way to define informativeness is with model uncertainty. The quality of the quantities that summarise model uncertainty (\eg, parameter uncertainty) can be used as selection criteria, where better active learning performance implies higher quality. However, its efficacy has been questioned as it conflates inference and model problems \cite{yao2019quality} and can under perform uniform sampling both from an empirical and theoretic perspective \cite{tifrea2022uniform}.

\section{Conclusion}
In this paper, we have argued for the importance of a principled and fundamental understanding of representing, learning and reasoning about uncertainty in NLG. We identified and organised the main sources of uncertainty, and highlighted the many important applications this perspective can power.

To do so, we laid down central concepts, their formal mathematical frameworks and the necessary vocabulary. We specifically drew attention to the possible worlds framework; probability as a way to express preference over these possible worlds; its two main interpretations; statistical tools to acquire and revise knowledge; and how these are commonly used in NLG. 

Then, building on the triangle of reference, we identified and organised the main sources of uncertainty in language production: input \textit{ambiguity}, \textit{errors}, and \textit{complexity}; the \textit{open-endedness of the communicative task}, the \textit{agent's personal perspective}, and the final \textit{linguistic realisation}--- and modelling: \textit{model specification}, \textit{parameter estimation}, and \textit{distribution shift}. We proposed a two-dimensional taxonomy to organise sources as a richer alternative to the aleatoric/epistemic distinction.

Finally, we highlighted exciting applications of disentangled representations of uncertainty in NLG. These span from applications related data uncertainty (decoding, controllable generation, explicit modelling of sub-populations), to model uncertainty (self-assessment, selective answering, OOD detection, active learning).

We hope to spark a shared understanding of uncertainty and inspire more principled and focused research in NLG. Crucially, we believe this perspective allows for systems that are more flexible, representative of the diversity of human language and its speakers, and reliable and trustworthy.

\section*{Acknowledgements}

This work was initiated at and benefited substantially from the Dagstuhl Research Meeting 22474: European Laboratory's for Learning and Intelligent Systems Third ELLIS NLP Workshop.

EI, CZ and WA are supported by the EU's Horizon Europe research and innovation programme (grant agreement No.\ 101070631, UTTER). CZ is also supported by
the Portuguese Recovery and Resilience Plan
through project C645008882-00000055 (NextGenAI, Center for Responsible AI).
ND has received funding by the German Federal Ministry of Education and Research and the Hessian Ministry of Higher Education, Research, Science and the Arts within their joint support of the National Research Center for Applied Cybersecurity ATHENE.
RF is supported by the European Research Council (ERC) Consolidator grant DREAM (No.\ 819455). BP is supported by ERC
Consolidator Grant DIALECT No.\ 101043235.
RS is supported by the Swiss National Science Foundation (project no.~176727).
JB is supported by the ELLIS Amsterdam Unit.

\bibliography{custom, anthology}
\bibliographystyle{acl_natbib}

\end{document}